%% file: main.tex
\documentclass[letterpaper, 10 pt, conference]{ieeeconf}  

\IEEEoverridecommandlockouts                              

\usepackage{graphics} 
\usepackage{epsfig} 
\usepackage{times}
\usepackage{epsfig}
\usepackage{graphicx}
\usepackage{amsmath}
\usepackage{amssymb}

\usepackage{soul}
\usepackage{xcolor}

\usepackage{algorithm} 
\usepackage{algpseudocode} 
\usepackage{multicol}
\usepackage{multirow}
\usepackage{booktabs}
\usepackage{caption} 
\usepackage{color,colortbl}
\definecolor{Gray}{gray}{0.9}

\title{\LARGE \bf
Fast Hierarchical Learning for Few-Shot Object Detection 
}

\author{Yihang She,  Goutam Bhat, Martin Danelljan, Fisher Yu \\Computer Vision Lab, ETH Zurich
}

\newcommand{\parsection}[1]{\vspace{0.1mm}\noindent\textbf{#1:}~}

\begin{document}

\maketitle
\thispagestyle{empty}
\pagestyle{empty}

\begin{abstract}

Transfer learning based approaches have recently achieved promising results on the few-shot detection task. These approaches however suffer from ``catastrophic forgetting'' issue due to finetuning of base detector, leading to sub-optimal performance on the base classes. Furthermore, the slow convergence rate of stochastic gradient descent (SGD) results in high latency and consequently restricts real-time applications. We tackle the aforementioned issues in this work. We pose few-shot detection as a hierarchical learning problem, where the novel classes are treated as the child classes of existing base classes and the background class. The detection heads for the novel classes are then trained using a specialized optimization strategy, leading to significantly lower training times compared to SGD. Our approach obtains competitive novel class performance on few-shot MS-COCO benchmark, while completely retaining the performance of the initial model on the base classes. We further demonstrate the application of our approach to a new class-refined few-shot detection task.

\end{abstract}
\input{01_introduction}
\input{02_related_work}

\input{03_method}
\input{04_experiments}

\input{05_conclusion}



{\small
\bibliographystyle{ieee.bst}
\bibliography{egbib}
}

\clearpage
\onecolumn
\input{supp}

\end{document}

%% file: 01_introduction.tex
\section{INTRODUCTION}

Few-shot object detection~\cite{kang2019few, zhang2021meta, wang2020frustratingly, fan2021generalized} is an important computer vision problem with practical applications in robotics. 
For instance, it can be used to deploy an autonomous agent in a new environment with \textit{unseen} objects, without having to collect large amount of training data. Alternatively a user may want a robot to detect new objects by showing just a few examples.
Few-shot object detection is an especially challenging problem since a model should learn to both classify an object and localize it using sparse data. This is further complicated in the generalized few-shot detection case~\cite{wang2020frustratingly}, where the model should retain the ability to detect a set of pre-learned base classes, while learning to detect novel classes.

One of the popular paradigms for the few-shot detection task is the use of transfer learning. These approaches~\cite{wang2020frustratingly,li2021class} aim to exploit general object detection knowledge learnt over a large dataset containing annotation for a set of base classes.
Here, a detection model is first trained on the data-abundant base classes. The final few layers of this model are then finetuned to jointly detect both the base and novel classes, using a few-shot dataset. While achieving promising results, especially on the novel classes, the transfer learning based methods suffer from two key limitations. Firstly, the finetuning of the base model on the few-shot dataset leads to a significant drop in the base class performance. This issue, termed as ``catastrophic forgetting'', is undesirable in practical applications where we may want a robot to detect new classes on the fly, while not forgetting the old knowledge. Secondly, the base model is finetuned using stochastic gradient descent, which takes long time to converge. This prohibits the use of the method for real-time applications.

\begin{figure}[t]
     \centering
     \includegraphics[width=\columnwidth]{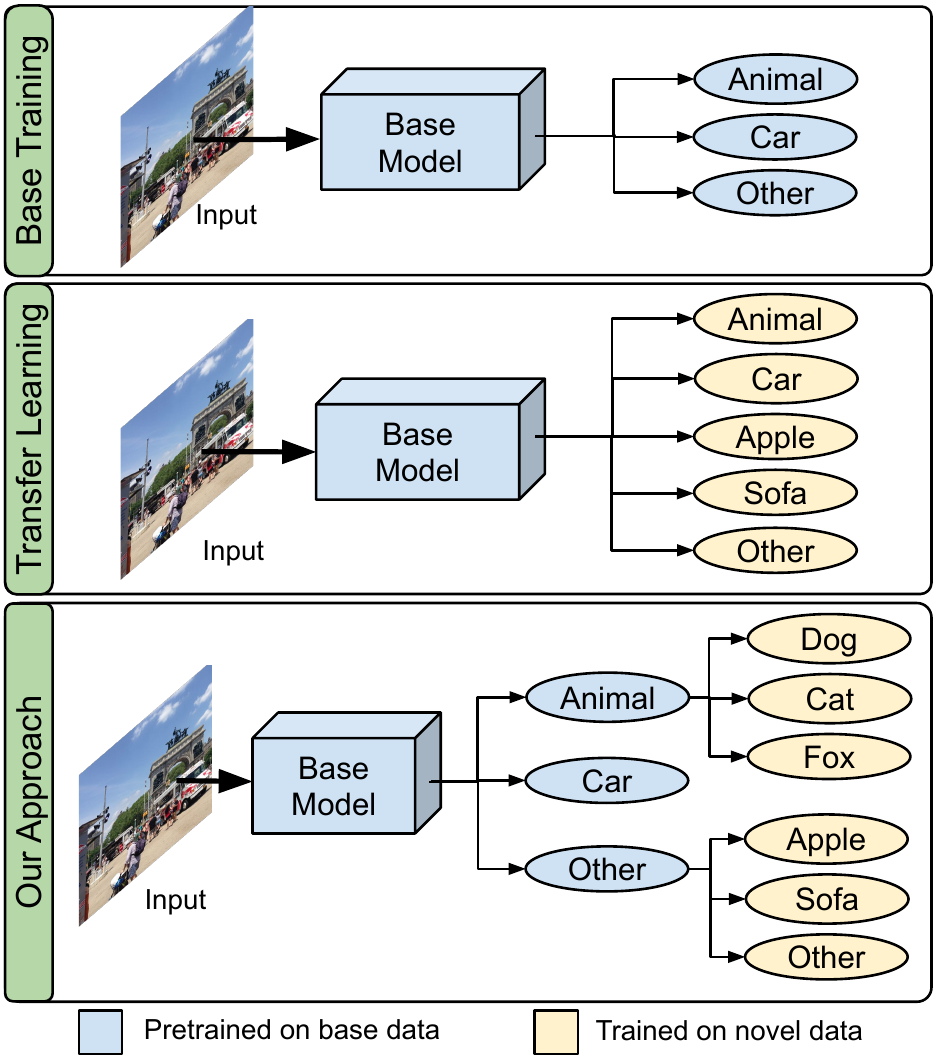}
    \caption{Transfer learning based approaches finetune a base detector to jointly detect both base and novel classes. However, this results in a drop in base class performance of the detector due to ``catastrophic forgetting''. Our approach instead detects novel classes in a hierarchical manner. This preserves the base class performance, while also enabling detecting child classes of existing base classes in a few-shot manner.}\vspace{-5mm}
    \label{fig:intro}
\end{figure}

In this work, we propose a novel few-shot detection approach to address the aforementioned issues. Our approach is based on the idea of posing few-shot detection as a hierarchical learning problem. We consider a general few-shot learning setting where we may wish to extend a detector to detect novel classes which are either a child class of an existing base class, or completely unrelated to the base classes. For example, given a model which detects ``animal'' and ``car'' classes, we may wish to additionally detect the animal types, e.g.\ ``cat'', ``dog'', or ``fox'', as well as unrelated novel classes ``apple'' and ``sofa''. To achieve this, we build a class hierarchy wherein the original base classes constitute a set of super-classes, which are then sub-divided into novel child classes. Specifically, the novel classes which are unrelated to any of the base classes are set as descendants of the ``other/background'' super-class. 
With such a hierarchy, we can first apply the base detector to detect the leaf base class objects (``car''), as well as the candidates for the base super-classes (``animal'' and ``other''). These candidates are then processed by separately trained novel predictors to detect the novel classes. See Figure~\ref{fig:intro} for an illustration.

Our hierarchical approach decouples the weights of the novel class predictors from the base detector. As a result, our approach retains the performance of the pre-trained detector on the base classes by design, addressing the ``catastrophic forgetting'' issue. 
Furthermore, we also introduce a specialized optimization strategy, based on the Newton's method, to speed up the learning of the novel predictors. By exploiting second-order information, our approach can adapt to detect the novel classes using only $30$ update steps. Consequently, our approach obtains over $10\times$ speed-up in computation time, compared to our transfer learning based baseline TFA~\cite{wang2020frustratingly}.

Our contributions can thus be summarized as follows:
\begin{itemize}
    \item We propose a simple yet effective hierarchical detection approach which completely alleviates the ``catastrophic forgetting'' on base classes, while obtaining competitive results on the novel classes.
    \item We present a Newton's method based optimization strategy which achieves mush faster convergence than traditional gradient descent.
    \item We introduce a new class-refined few-shot detection task where a method should also be able to learn fine-grained classification for existing base classes.
\end{itemize}

%% file: 02_related_work.tex
\section{RELATED WORK}
\parsection{Few-Shot Object Detection}
Existing literature mainly adopt two paradigms to tackle the few-shot object detection problem: meta learning-based approach \cite{kang2019few, perez2020incremental, fan2020few, zhang2021meta} and transfer learning-based approach \cite{chen2018lstd, wang2020frustratingly, li2021class, fan2021generalized, qiao2021defrcn}. For meta learning-based approach, researchers leverage the meta-learned task-level knowledge to the detection task with limited training data. MetaYOLO \cite{kang2019few} meta learned a feature learner module to extract the generic features of novel objects and a reweighting module to make predictions provided these features. 
Fan et al.\ \cite{fan2020few} proposed Attention-RPN and Multi-Relation Detector to learn a metric space to measure the similarity of object pairs for detection. Meta-DETR \cite{zhang2021meta} meta learned an encoder-decoder transformer for the few-shot detection. 

For transfer learning-based approach, LSTD \cite{chen2018lstd} is one of the early works that adapted the detector learned on data-abundant objects to the target domain of few-shot novel objects. Wang et al.\ \cite{wang2020frustratingly} proposed the two-stage fine-tuning approach TFA. In the first stage, a base predictor was trained for data-abundant base objects. The final layers of the detector were then tuned in the second stage, on a balanced few-shot dataset containing both base and novel classes. This tuning-based approach is simple yet effective, and outperformed previous methods using meta-learning. 
Compared to TFA, LEAST \cite{li2021class} fine-tuned more layers on novel classes, leading to a better novel class performance, albeit with a deterioration on the base class performance. To mitigate this catastrophic forgetting, they further applied knowledge distillation and the clustered exemplars of base objects. 
DeFRCN \cite{qiao2021defrcn} fine-tuned the entire detector of Faster R-CNN by jointly training it with two auxiliary modules 
to improve novel class performance.
Fan et al.\ \cite{fan2021generalized} proposed Retentive R-CNN, which inherited the tuning approach of TFA with an auxiliary consistency loss to distill the knowledge of the base detector. Retentive R-CNN achieved competitive performance on novel classes, while maintaining the performance of the pre-trained detector on base classes. 
In this paper, we propose an alternate hierarchical detection approach which can achieve similar results to Retentive R-CNN, while being much simpler and general. 

\parsection{Incremental Learning and Refined Classification}
Incremental learning aims to incrementally learn new knowledge from a stream of data while preserving its previous knowledge \cite{li2017learning, riemer2018learning, hu2018overcoming,rebuffi2017icarl, tao2020few, xiao2014error, zhang2021few, bormann2021real}.
A real-world scenario which is often neglected is that over time, humans learn not only new entities, but also refined granularity of previously learned entities. 
Abdelsalam et al.\ \cite{abdelsalam2021iirc} propose the Incremental Learning and Refined Classification (IIRC) setup related to this scenario. Here, each class has two granularity levels of labels to simulate the process of incremental learning from coarse-grained categories to fine-grained categories. Following the IIRC setup, Wang et al.\ \cite{wang2021hcv} proposed HCV to learn the fine-grained categories while retaining previous knowledge. HCV aims to identify hierarchical relationship between classes and exploit this knowledge for the IIRC task. 

\parsection{Hierarchy for few-shot learning}
Li et al.\ \cite{li2019large} perform large-scale few-shot learning by using class hierarchy which encodes semantic relations between base and novel classes. The prior knowledge from class hierarchy is used to learn transferrable visual features. Liu et al.\ \cite{liu2020many} use class hierarchy to perform coarse-to-fine classification. In contrast to these works, we show that the idea of hierarchy can be effectively used to address the ``catastrophic forgetting'' issue in few-shot detection.


\parsection{Optimization Methods for Few-Shot Learning}
Bertinetto et al.\ \cite{bertinetto2018meta} noted that updating only the parameters sensitive to specific classes for few-shot classification task leads to a shallow learning problem. This enables developing adaptation strategies that are more efficient than standard gradient descent. Consequently, they proposed ridge and sigmoid regression based classifiers with closed-form solutions to achieve fast convergence for the meta-learning-based few-shot classification. Lee et al.\ \cite{lee2019meta} meta-learn representations for few-shot classification using discriminative linear classifiers. Several works have utilized the steepest-descent optimization strategy to train shallow learners for tackling few-shot learning problem arising in object tracking \cite{danelljan2016beyond, sun2018correlation, bhat2019learning}, video object segmentation \cite{bhat2020learning} and classification \cite{tripathi2020few}. 
A few works~\cite{alamri2019contextual, alamri2020improving} have employed conjugate gradient (CG) as a black box optimization tool for object detection. In this work, we develop a specialized optimization strategy based on CG to perform efficient few-shot detection. By running
extensive experiments, we show that our optimization approach obtains similar performance to SGD while being much faster.


%% file: 03_method.tex
\section{METHOD}
In this work, we propose a few-shot detection approach that can learn to efficiently detect novel classes, while fully retaining the performance of the original detector on the base classes. This is achieved by i) introducing a hierarchical detection approach which preserves the performance on the base classes by design, while obtaining competitive results on the novel classes, and ii) utilizing a specialized optimization approach which leads to faster model adaptation on novel classes. Our approach is detailed in subsequent sections. 

\subsection{Problem statement}
We tackle the generalized few-shot learning setting employed in previous works~\cite{wang2020frustratingly, fan2021generalized}. Here, a method is given a large base dataset $D_b$ containing annotated samples for a set of base object classes $C_b$, which can be used to learn a base detection model $M_b$. Next, given a small dataset $D_n$ for a set of novel classes $C_n$, the goal is to adapt the base model $M_b$ to detect the novel classes $C_n$, in addition to the original base classes $C_b$. The novel dataset $D_n$ is assumed to contain only $K$ examples ($K < 30$) per class. Furthermore, the method is only allowed to access a small $K$-shot subset $D'_b$ of the base dataset $D_b$ when adapting the base model to detect the novel classes. Thus the method should be able to easily adapt to detect the novel classes $C_n$ using a small dataset $D_n \cup D'_b $, while still retaining the ability to detect the base classes $C_b$. Furthermore, for practical applications in e.g.\ robotics, the adaptation to novel classes is expected to be fast in order to ensure real-time performance.

\subsection{Motivation}
We base our approach on the recently introduced Two-stage Fine-tuning Approach (TFA) \cite{wang2020frustratingly}. TFA is a transfer learning approach for few-shot object detection which has obtained promising results. TFA employs the Faster-RCNN~\cite{ren2015faster} as the detector architecture. Faster-RCNN consists of a convolutional neural network (CNN) module for extracting generic image features, a Regional Proposal Network (RPN) to generate proposals for potential objects, a Region of Interest (ROI) feature extractor to compute features from the sampled proposals, and a predictor head to output the detections, given the ROI features. The predictor $\mathcal{P} = \{\mathcal{C}, \mathcal{R}\}$ consists of two separate linear layers: a classifier $\mathcal{C}$ to predict object class for each proposal and a bounding box regressor $\mathcal{R}$ to localize each proposal.

TFA proposes to first train a Faster-RCNN model on the data-abundant $D_b$ to obtain a base detector $M_b$, with predictor $\mathcal{P}_b$. Next, when provided the novel dataset $D_n$ in the second stage, TFA extends the predictor $\mathcal{P}_b$ to also output detections for the novel classes. This extended predictor, denoted as $\mathcal{P}_n$, is then fine-tuned on the combined dataset $D_n \cup D'_b$ by minimizing a loss $\mathcal{L} = \mathcal{L}_{cls} + \mathcal{L}_\text{loc}$ using the stochastic gradient descent optimizer. Here, $\mathcal{L}_\text{cls}$ is the cross entropy loss for classifier $\mathcal{C}$ while $\mathcal{L}_\text{loc}$ is the smooth $L_1$ loss for box regressor $\mathcal{R}$. We refer to the detector using the finetuned predictor $\mathcal{P}_n$ as $M_n$.

The two-stage training strategy allows TFA to leverage the strong backbone feature extractor and the RPN modules trained on the larger base dataset $D_b$ to obtain improved performance on the data-scarce novel categories $C_n$. 
However, it suffers from two significant issues which limits its applicability to practical applications. 


\begin{table}[t]
\centering
\begin{tabular}{@{}l|c|ccccc@{}}
\toprule
~& Base & 1-shot & 2-shot & 3-shot & 5-shot & 10-shot\\ \midrule
bAP  & 39.2 & 34.1 & 34.7 & 34.7 & 34.7 & 35.0 \\ \bottomrule
\end{tabular}\vspace{-0mm}
\caption{Average precision of TFA~\cite{wang2020frustratingly} on base classes (bAP) over different shots on MS-COCO dataset~\cite{lin2014microsoft}. TFA suffers from a significant drop in bAP, compared to the pre-trained base model, due to ``catastrophic forgetting''.}\vspace{-5mm}
\label{tab:forgetting}
\end{table}

\begin{figure*}[t]
     \centering
     \includegraphics[width=\textwidth]{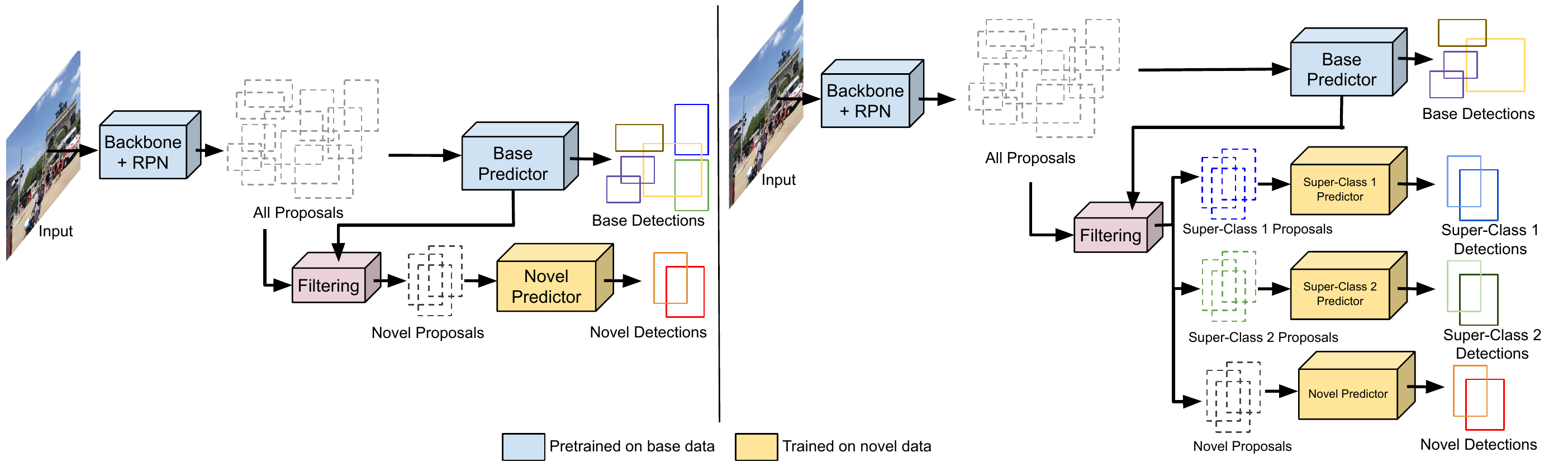}
    \caption{\textbf{Left:} An overview of our hierarchical approach for generalized few-shot detection. Given object proposals, we use a base predictor trained on the large base dataset to obtain detections for the base classes. Next, we filter the proposals which are classified as ``other objects'' or ``background'' and pass them through a novel predictor to perform novel class detection. \textbf{Right:} An extension of our approach to perform class-refined detection. In addition to detecting novel classes which are unrelated to the base classes, we also learn to detect child classes of existing base classes using few-shot dataset.}\vspace{-5mm}
    \label{fig:method_std}
\end{figure*}

\parsection{Catastrophic forgetting} In the second stage, TFA finetunes the predictor on a small balanced dataset containing both novel and base categories to obtain the detector $M_n$. This finetuning can lead to a significant drop in the base class performance, compared to the base detector $M_b$ which was trained on a much larger dataset $D_b$. This ``catastrophic forgetting''  problem is illustrated in Tab.~\ref{tab:forgetting}. Compared to the pretrained base detector $M_b$, the finetuned detector $M_n$ obtains much lower average precision score on the base classes (bAP), even in the 10-shot case (39.2 vs 35.0). This is undesirable in cases when the performance on base classes is equally important as the performance on novel classes.




\parsection{Slow convergence} TFA uses the Stochastic Gradient Descent (SGD) to finetune the predictor $\mathcal{P}_n$ in the second stage. While SGD is computationally cheap for each iteration, it suffers from slow convergence. Thus, a large number of SGD iterations is required to adapt the base detector $M_b$ for novel classes, leading to high computational times. 

In this work, we address the aforementioned issues with TFA by proposing a novel few-shot detection framework.

\subsection{Hierarchical Detection Approach} \label{sec: hierarchical detection approach}


Here, we present our Hierarchical Detection Approach (HDA) for generalized few-shot detection (see Fig.\ \ref{fig:method_std}). 
We note that the base model $M_b$ is pre-trained on a large dataset $D_b$ containing abundant examples of base classes $C_b$.
Thus, the detector $M_b$ should already achieve high detection performance on the base classes. Finetuning it further on a smaller subset $D_n \cup D'_b$, as in TFA, is likely to only reduce the base class performance due to overfitting.
Furthermore, the base dataset $D_b$ also contains a large number of background objects not belonging to $C_b$. Consequently, the base detector $M_b$ should be able to classify most of the unseen object classes, including the novel classes, as background. Under these settings, we can pose generalized few-shot detection as a hierarchical detection problem, as described next. 



Similar to TFA, we first train a Faster-RCNN base detector $M_b$ to detect the base classes $C_b$ using the large-scale base dataset $D_b$. 
Next, instead of finetuning the predictor $\mathcal{P}_b$ in order to adapt the model for novel classes, we employ an alternate hierarchical approach. 
We first apply the detector to generate object proposals and use the base predictor $\mathcal{P}_b$ to obtain the classification scores and refined boxes for each of these proposals. Next, we perform the standard post-processing steps employed in Faster-RCNN to obtain the predictions for the base classes $C_b$. As discussed before, this should provide high quality base class detections, since the base detector $M_b$ is trained on a large dataset.


Next, among the proposals which were not included in the base detections, we first select the proposals which are classified as ``other objects/background''. That is, we select the proposals for which the classification score predicted by $\mathcal{P}_b$ is highest for the background class. Note that these filtered proposals include both the novel class $C_n$ proposals, as well as proposals for other unseen classes and the background. In order to obtain the novel class detections from these proposals, we use a $\mathcal{P}_n$, trained to detect only the novel classes $C_n$. The novel predictor $\mathcal{P}_n$ can be easily trained using the novel dataset $D_n$, or the combined dataset $D_n \cup D'_b$. 
By completely decoupling the training of the novel predictor $\mathcal{P}_n$ from the pre-trained base predictor $\mathcal{P}_b$,
our approach completely retains the performance of the pre-trained detector $D_b$ on the base classes. Furthermore, our novel predictor $\mathcal{P}_n$ is trained to only detect the novel classes $C_n$, rather than the combined novel and base classes as in TFA. This simplifies the training of the novel predictor.

\subsection{Class-Refined Hierarchical Detection} 
\label{sec:cls_refined}
In our previous discussion, we assume that the base detector $M_b$ can reliably classify the novel classes $C_n$ as background. This is generally the case when novel classes are semantically distinct from the base classes. However in many applications, we may wish to extend a base detector to detect child classes of an existing base class in a few-shot manner, in addition to detect completely novel classes. 
For instance, given a detector which can detect the ``animal'' class, we may want to extend it to detect fine-grained classes of ``cat'' and ``dog'' using only a few examples. 
In such cases, the base detector is likely to classify instances of cats and dogs as ``animal'', instead of background. However, our approach can be easily modified to perform such class-refined few-shot detection, as described next. 


Let $C_b = \{b_i\}_{i=1}^{B+1}$ denote the set of base classes and the background class denoted as $b_{B+1}$. Furthermore, $C_n = \{n_i\}_{i=1}^N$ denotes the set of novel classes, some of which can be semantically related to an existing base class. We represent this relationship by dividing the set of novel classes into disjoint subsets $C^i_n = \{c_j\}_{j=1}^{N_i}$, $i \in \{1, 2, \dots, B+1\}$, where some of the sets $C^i_n$ can be empty. Here, the subset $C^i_n$ contains the set of novel classes which are semantically related to the base class $b_i$. Note that such a sub-division can be set manually, using e.g.\ WordNet~\cite{miller1998wordnet} hierarchy. Alternatively, it can also be obtained automatically using the output of the base detector $M_b$ on the novel dataset $D_n$. For instance, if the base detector frequently classifies a novel class $n_j$ as the base class $b_i$, then the novel class $n_j$ can be assigned to the subset $C^i_n$. 

Given the sub-division $C^i_n$ of the novel classes, we can use a similar hierarchical detection strategy as described in the previous section to perform few-shot detection. Our approach is visualized in Fig.\ \ref{fig:method_std}. We first apply the base detector $M_b$ to obtain the object proposals, classification scores, and refined boxes for each of the proposals. Next, for each base class $b_i$, we select the proposals for which the classification score corresponding to that class is the highest. Note that these filtered proposals correspond to both the base class $b_i$, as well as the set of novel classes $C^i_n$ which are child classes of $b_i$.
In order to distinguish between them, we train a novel predictor $\mathcal{P}^i_n$ separately for each subset $C^i_n$. The predictor is trained to classify the proposal into either the base class $b_i$, or one of the child classes belonging to $C^i_n$. Finally, the standard post-processing steps, e.g.\ NMS are applied to the output of the predictor $\mathcal{P}^i_n$ to generate the final detections. 

\begin{table*}[t] 
\centering
\resizebox{\textwidth}{!}{
    \begin{tabular}{l|ccccc|ccccc|ccccc}
    \toprule
    \multirow{2}*{Model} & \multicolumn{5}{c|}{AP} & \multicolumn{5}{c|}{bAP} & \multicolumn{5}{c}{nAP} \\ 
    \cmidrule(lr){2-6} \cmidrule(lr){7-11} \cmidrule(lr){12-16}
    ~& 1 & 2 & 3 & 5 & 10 & 1 & 2 & 3 & 5 & 10 & 1 & 2 & 3 & 5 & 10 \\ 
    \midrule
   FRCN+ft-full\cite{wang2020frustratingly}           & 16.2 & 15.8 & 15.0 & 14.4 & 13.4 & 21.0 & 20.0 & 18.8 & 17.6 & 16.1 & 1.7 & 3.1 & 3.7 & 4.6 & 5.5  \\
    TFA w/fc\cite{wang2020frustratingly}               & 24.0 & 24.5 & 24.9 & 25.6 & 26.2 & 31.5 & 31.4 & 31.5 & 31.8 & 32.0 & 1.6 & 3.8 & 5.0 & 6.9 & 9.1  \\
    TFA w/cos\cite{wang2020frustratingly}              & 24.4 & 24.9 & 25.3 & 25.9 & 26.6 & 31.9 & 31.9 & 32.0 & 32.3 & 32.4 & 1.9 & 3.9 & 5.1 & 7.0 & 9.1  \\
    LEAST\cite{li2021class}                  & - & - & - & - & - & 29.5 & - & - & 31.3 & 31.3 & \textcolor{blue}{4.2} & - & - & \textcolor{blue}{9.3} & \textcolor{blue}{12.8}  \\
    DeFRCN\cite{qiao2021defrcn} & 24.4 & 25.7 & 26.6 & 27.8 & 29.7 & 30.4 & 31.4 & 32.1 & 32.6 & 34.0 & \textcolor{red}{4.8} & \textcolor{red}{8.5} & \textcolor{red}{10.7} & \textcolor{red}{13.6} & \textcolor{red}{16.8} \\
    Retentive R-CNN\cite{fan2021generalized}        & - & - & - & \textcolor{red}{31.4} & \textcolor{red}{31.8} & - & - & - & \textcolor{red}{39.3} & \textcolor{red}{39.2} & - & - & - & 7.7 & 9.5  \\
    \midrule
    \rowcolor{Gray} 
    HDA      & \textcolor{red}{30.2} & \textcolor{red}{30.6} & \textcolor{red}{31.0} & \textcolor{blue}{31.2} & \textcolor{red}{31.8} & \textcolor{red}{39.2} & \textcolor{red}{39.2} & \textcolor{red}{39.2} & \textcolor{blue}{39.2} & \textcolor{red}{39.2} & 3.0 & \textcolor{blue}{4.7} & \textcolor{blue}{5.6} & 7.1 & 9.1  \\ 
    \rowcolor{Gray}
    HDA-wo-Aug          & \textcolor{blue}{30.0} & \textcolor{blue}{30.4} & \textcolor{blue}{30.7} & 31.0 & \textcolor{blue}{31.4} & 39.2 & 39.2 & 39.2 & \textcolor{blue}{39.2} & \textcolor{red}{39.2} & 2.4 & 4.1 & 5.1 & 6.4 & 8.0  \\
    TFA-Fast          & 24.2 & 25.3 & 26.2 & 27.3 & 28.5 & 31.4 & 32.4 & 33.3 & 34.3 & \textcolor{blue}{35.2} & 2.6 & 4.0 & 4.9 & 6.4 & 8.5  \\
    TFA$^*$    & 24.1 & 24.7 & 25.0 & 25.7 & 26.1 & 31.5 & 31.7 & 31.6 & 31.8 & 31.6 & 2.3 & 3.8 & 4.9 & 6.4 & 8.3  \\
    \bottomrule
    \end{tabular}
}\vspace{-1mm}
\caption{Generalized few-shot detection performance on MS-COCO~\cite{lin2014microsoft} dataset. We report the combined average precision (AP) score, as well as the performance over the base classes (bAP) and novel classes (nAP) for 1, 2, 3, 5, and 10-shot datasets. The top half provides comparison with recent approaches, while the bottom half provides an ablation comparison.}\vspace{-4mm}
\label{tab:sota_std}
\end{table*}

\subsection{Fast Optimization for Few-Shot Detection} \label{sec: fast optimization for few-shot detection}
Here, we describe our fast optimization approach based on Newton's method used to efficiently train the novel predictor $\mathcal{P}_n$. Recall that our novel predictor only consists of two separate linear layers $\mathcal{C}$ and $\mathcal{R}$ for classification and bounding box regression, respectively. As a result, training the novel predictor constitutes a shallow learning problem. This enables using specialized optimization techniques to obtain fast convergence when learning the novel predictor.


In order to design our optimization strategy, we first replace the bounding box regression loss $\mathcal{L}_\text{loc}$ from smooth-$L_1$ loss with a $L_2$ loss. This, coupled with the convexity of the cross-entropy loss employed as the classification loss $\mathcal{L}_\text{cls}$ provides us a convex and differentiable objective $\mathcal{L} = \mathcal{L}_\text{cls} + \mathcal{L}_\text{loc}$. 
Our goal is then to find the optimal weights $w^*$ which minimizes the objective $\mathcal{L}(w)$, where $w$ denote the set of trainable weights in the predictor $\mathcal{P}_n$.
One well-known technique to minimize such a formulation is Newton's method, which leverages the second-order information to iteratively find the optimal solution. In each iteration, Newton's method first obtains a quadratic approximation $\Tilde{\mathcal{L}}_{w_i}(\Delta{w}) \approx \mathcal{L}(w_i + \Delta{w})$ of the objective at the current estimate $w_i$, 
\begin{equation}
\label{eq:l_quad_approx}
    \Tilde{\mathcal{L}}_{w_i}(\Delta{w})\approx \mathcal{L}(w_i)+ \Delta w_i^T J_{w_i}+\frac{1}{2}\Delta w_i^T H_{w_i} \Delta w_i
\end{equation}
Here, $J_{w_i}$ and $H_{w_i}$ denote the Jacobian and Hessian, respectively, of the objective $\mathcal{L}(w)$ at $w_i$. The next update $\Delta w_i$ for the weights $w$ is then obtained by minimizing $\Tilde{\mathcal{L}}_{w_i}(\Delta{w})$ w.r.t.\ $\Delta{w}$. This provides a closed-form expression for the update,  
\begin{equation}
\label{eq:update_step}
    \Delta w_i = -H_{w_i}^{-1}\cdot J_{w_i}
\end{equation}
In practice, calculating the inverse of Hessian matrix $H_{w_i}^{-1}$ is computationally expensive and can be numerically unstable. 
However, thanks to the use of cross entropy and least squares loss in our objective $\mathcal{L}$, the Hessian matrix $H_{w_i}$ is positive-definite. As a result, instead of explicitly computing the inverse of the Hessian, we can obtain the step $\Delta w_i$ as the solution of the linear system $H_{w_i} \Delta w_i = -J_{w_i}$. We use the iterative Conjugate Gradient (CG) algorithm for this purpose. We refer to~\cite{shewchuk1994introduction} for a detailed description of the CG algorithm. For efficiency, we only compute an approximate solution of the linear system by running $N_\text{CG}$ Conjugate Gradient steps in each Newton iteration.

%% file: 04_experiments.tex
\section{EXPERIMENTS} \label{sec: experiments}
We compare our approach with existing methods and provide a detailed analysis of our approach.

\subsection{Experimental Settings} \label{sec: experimental settings}

We evaluate our approach on the generalized few-shot detection (G-FSD) benchmark based on the MS-COCO~\cite{lin2014microsoft} dataset, introduced in~\cite{wang2020frustratingly}. The benchmark uses the same class splits as in previous few-shot detection works~\cite{wang2020frustratingly, li2021class,fan2021generalized}. We report results on 1, 2, 3, 5, and 10-shot datasets. For each $K$-shot setting, the benchmark samples 10 random datasets, and computes the mean scores over them in order to obtain a robust evaluation metric. 
For each method, we report the combined average precision scores over all classes (AP), as well as the average precision over the base (bAP) and novel (nAP) classes, separately. This allows us to analyse the capability of the model to quickly learn on novel classes, as well as its ability to retain knowledge on the base classes.

\begin{table}[t] 
\centering
\resizebox{0.5\textwidth}{!}{
    \begin{tabular}{l|ccccc}
    \toprule
    \multirow{2}*{Model} & \multicolumn{5}{c}{Runtime (seconds)} \\
    \cmidrule(lr){2-6} 
    ~& 1 & 2 & 3 & 5 & 10 \\ \midrule
    \rowcolor{Gray}
    HDA   & 10.19 & 19.84 & 28.18 & 45.74 & 94.18 \\
    \rowcolor{Gray}
    HDA-wo-Aug       & 10.32 & 19.15 & 28.76 & 46.74 & 94.19 \\
    TFA-Fast     & 13.14 & 24.30 & 33.44 & 54.36 & 107.81  \\
    TFA$^*$    & 121.53 & 245.46 & 364.83 & 617.83 & 1245.94 \\ \bottomrule
    \end{tabular}
}\vspace{-2mm}
\caption{Comparison of total training time on the few-shot dataset for the baseline TFA$^*$ and different variants of our approach over 1, 2, 3, 5, and 10 shot datasets.}\vspace{-4mm}
\label{tab:runtime}
\end{table}

\begin{figure*}[ht]
     \centering
         \includegraphics[width=\textwidth]{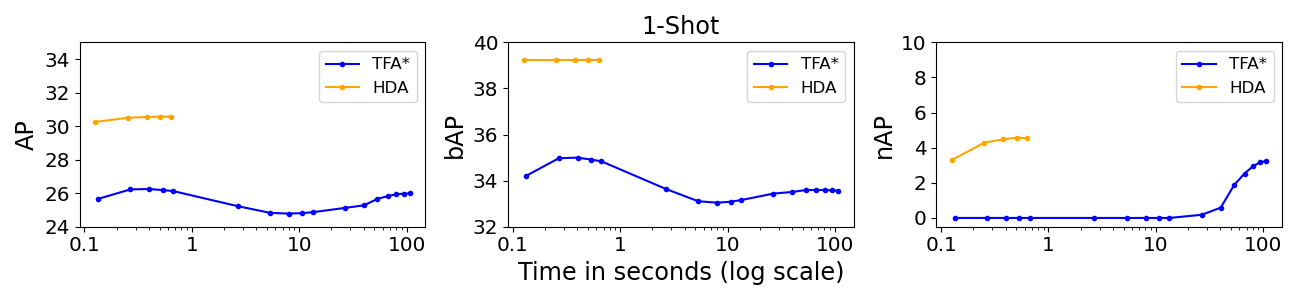}
     \vspace{-7mm}
    \caption{Convergence plots showing the model performance for 1-shot detection, in terms of AP, bAP, and nAP, for different training times (in seconds). We exclude the initial time for feature extraction ($\sim$10 seconds for both methods). Note that the x-axis follows logarithmic scale. Compared to the baseline TFA$^*$, our approach obtains around $100$ times faster convergence in nAP score, while preserving the bAP of the base model.}\vspace{-5mm}
    \label{fig:convergence_plot}
\end{figure*}

\parsection{Implementation Details}
We use Faster R-CNN \cite{ren2015faster} as the detector architecture with ResNet-101 \cite{he2016deep} with FPN \cite{lin2017feature} as backbone. As our base model $M_b$, we use the pre-trained model from TFA~\cite{wang2020frustratingly} which has been trained on the base dataset $D_b$.
The novel class predictor $\mathcal{P}_n$ is trained to detect the novel classes using the proposals classified as ``other/background'' by $M_b$.
For training $\mathcal{P}_n$, we initialize the weights randomly using a zero-mean normal distribution with standard deviation $0.01$.
The predictor is trained for 30 Newton iterations, with $N_\text{CG} = 2$ CG iterations for every single Newton iteration. For our SGD based baselines, we follow the hyper-parameters used in TFA. 
Since the size of our few-shot dataset is small and only the predictor layer is updated during our training, we first extract the proposals and box features for every training image and store it in memory. The predictor is then trained directly on the extracted features, thus avoiding the re-computation of these features and speeding up the training. We perform simple data augmentation on the extracted box features in order to mitigate overfitting.
For each Newton iteration, we first make 5 copies of each box features. Next, we apply zero-mean Gaussian noise with standard deviation $0.1$ and perform dropout with a rate of $0.5$. 

\subsection{Comparison Experiments}
\label{exp:comparion}

Here, we compare our method HDA with the existing few-shot detection approaches. We compare with a baseline  
\texttt{FRCN+ft-full} that fine-tunes all layers of the base Faster-RCNN detector on novel classes, as well as variants of the baseline TFA~\cite{wang2020frustratingly}, including \texttt{TFA w/fc} with the fully-connected layer classifier and \texttt{TFA w/cos} with the cosine similarity-based classifier. We also compare with recent transfer learning based approaches \texttt{LEAST} \cite{li2021class}, \texttt{Retentive R-CNN} \cite{fan2021generalized} and \texttt{DeFRCN}~\cite{qiao2021defrcn}. 

Table~\ref{tab:sota_std} reports the mean average precision (AP, bAP, and nAP) of previous approaches and our method on different shots. Compared to the baseline TFA, our approach obtains consistent improvements in novel class performance (nAP), while being significantly better on the base classes.
DeFRCN achieves the best performance on the novel classes while suffering from a significant drop on the base classes. Similarly, LEAST obtains the second best performance on the novel classes in the 1, 5, and 10 shot settings, thanks to its aggressive fine-tuning strategy. However, this results in a significant drop in the base class performance due to catastrophic forgetting. In contrast, our approach obtains competitive results on the novel classes, while fully retaining the performance of the base model on the base classes. Consequently, our approach obtains the best AP score over all classes, on par with the recent Retentive R-CNN. Note that this is achieved while being computationally fast, 
thereby making our approach more suitable for practical applications.  


\subsection{Ablation Study}
\label{exp:ablation}
In this section, we analyze the impact of our contributions by evaluating different variants of our approach. We start with a variant of the baseline TFA, denoted as, \texttt{TFA$^*$}, wherein we replace smooth-L1 loss for bounding box regression with L2 loss. Furthermore, we run SGD directly on the extracted box features as in our approach to obtain a fair runtime comparison. Next, we replace the SGD-based finetuning in TFA with our fast Newton's method based optimization strategy to obtain \texttt{TFA-Fast}. We also evaluate a variant of our approach \texttt{HDA-wo-Aug} without the data augmentation strategy (Sec.~\ref{sec: experimental settings}) to analyze its impact.

The results of this comparison, in terms of AP, bAP, and nAP is provided in Table~\ref{tab:sota_std}. The total training times on the few-shot datasets for each of the methods is additionally provided in Table~\ref{tab:runtime}. Note that the runtime for all methods are obtained using a single TITAN Xp GPU. Replacing the SGD optimizer in \texttt{TFA$^*$} with our fast optimizer leads to $\sim$ $10 \times$ speed-up in computation time, while also improving the AP score.
Using our hierarchical detection approach instead of two-stage fine-tuning addresses the ``catastrophic forgetting'' issue in \texttt{TFA+Fast}, leading to a significant $7.8$ improvement in bAP while achieving a similar nAP score on the 1-shot dataset. Additionally employing data augmentation for training the novel predictor $\mathcal{P}_n$ improves the novel class performance by $0.7$ nAP in the $5$-shot setting. 

We also provide convergence plots showing the model performance on 1-shot dataset for HDA and TFA$^*$, in terms of AP, bAP, and nAP, for different training times in Fig.\ \ref{fig:convergence_plot}. Note that we exclude the initial time for feature extraction (around 10 seconds for both methods) 
and only show the time for model optimization. Our approach obtains around $100$ times faster convergence in nAP score compared to the baseline TFA$^*$. 
Furthermore, our approach fully preserves the base class performance (bAP) of the base model. In contrast, the bAP for baseline TFA$^*$ fluctuates and is significantly lower compared to our approach due to catastrophic forgetting.

\begin{figure*}[t]
     \centering
     \includegraphics[width=\textwidth]{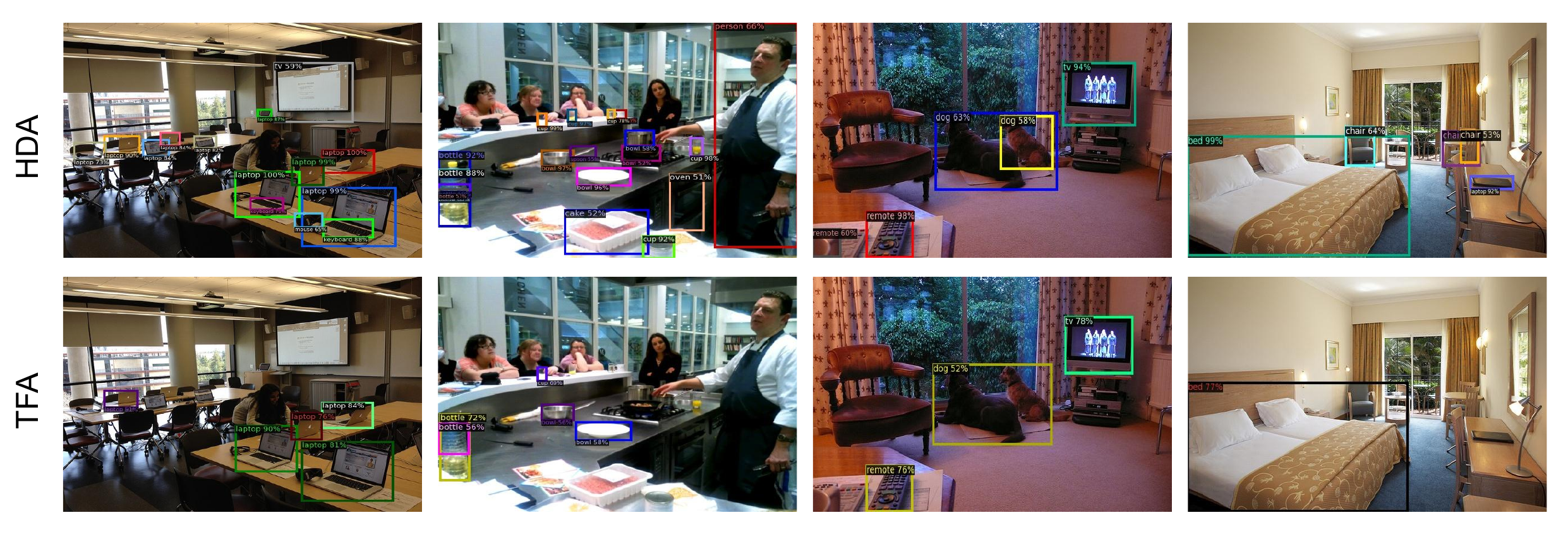}\vspace{-4mm}
    \caption{Qualitative comparison with baseline TFA on test images from 10-shot MS-COCO dataset. Our approach provides better detection performance, compared to TFA, over both the base (laptop, remote) as well as novel classes (TV, person).}\vspace{-3mm}
    \label{fig:visual_plot}
\end{figure*}

\begin{table*}[t] 
\centering
\resizebox{\textwidth}{!}{
    \begin{tabular}{l|ccc|ccc|ccc|ccc|ccc}
    \toprule
    \multirow{2}*{Model} & \multicolumn{3}{c|}{AP} & \multicolumn{3}{c|}{bAP} & \multicolumn{3}{c|}{aAP} & \multicolumn{3}{c|}{fAP} & \multicolumn{3}{c}{nAP} \\ 
    \cmidrule(lr){2-4} \cmidrule(lr){5-7} \cmidrule(lr){8-10} \cmidrule(lr){11-13} \cmidrule(lr){14-16}
    ~& 1 & 5 & 10 & 1 & 5 & 10 & 1 & 5 & 10 & 1 & 5 & 10 & 1 & 5 & 10  \\ 
    \midrule
    \rowcolor{Gray}
    HDA & \textcolor{red}{21.2} & \textcolor{red}{22.4} & \textcolor{red}{22.9} & \textcolor{red}{39.2} & \textcolor{red}{39.2} & \textcolor{red}{39.2} & 6.9 & 10.5 & 12.0 & 4.0 & 6.4 & 7.6 & \textcolor{red}{0.9} & \textcolor{red}{2.7} & \textcolor{red}{3.4}      \\ 
    \rowcolor{Gray}
    HDA-wo-Aug & \textcolor{red}{21.2} & \textcolor{red}{22.4} & \textcolor{red}{22.9} & \textcolor{red}{39.2} & \textcolor{red}{39.2} & \textcolor{red}{39.2} & 7.3 & 10.7 & 12.2 & 4.1 & 6.6 & 7.5 & 0.8 & 2.6 & \textcolor{red}{3.4}          \\
    TFA & 16.9 & 18.4 & 19.2 & 30.6 & 30.9 & 31.2 & \textcolor{red}{7.7} & \textcolor{red}{12.3} & \textcolor{red}{14.2} & \textcolor{red}{4.7} & \textcolor{red}{7.5} & \textcolor{red}{8.4} & 0.4 & 2.2 & 3.1     \\
    \bottomrule
    \end{tabular}
}\vspace{-1mm}
\caption{Comparison with TFA on the class-refined few-shot detection in terms of overall AP, performance on novel (nAP) and base (bAP) classes, as well as the refined detection performance over ``animal'' (aAP) and ``food'' (fAP) super-classes.}\vspace{-5mm} 
\label{tab:refined_fsdet}
\end{table*}

\parsection{Qualitative Comparison} Figure~\ref{fig:visual_plot} provides a qualitative comparison of our approach with baseline TFA on MS-COCO test set. Both models are trained on the 10-shot dataset. 
Note that the images were selected randomly, while only ensuring that they had multiple diverse objects.
Detections that have confidence score greater than 0.5 are displayed. 
Images in the first two columns are dominated by base classes (e.g.\ laptop, oven, bottle, cup).
As can be seen in the output detection, our approach detects greater number of base class instances compared to TFA, while being able to detect novel objects as well (TV and person). Similarly, in the images dominated by the novel classes e.g.\ dog, chair (last two columns), our approach provides similar or better results compared to TFA. 

\subsection{Class-Refined Few-Shot Detection}
\label{exp:cls_refined}
In Sec.~\ref{sec:cls_refined}, we present an extension of our hierarchical detection approach which can be used to detect novel classes that are semantically related to existing base classes in a few-shot manner. Here, we evaluate this approach on a new class-refined few-shot detection setting. We consider a practical scenario where the goal is to extend a base detector to either i) detect novel classes unrelated to the base classes, or ii) detect child classes for an existing base class using only a few samples. For example, given a base detector trained to detect ``animal'' and ``vehicle'' classes, we may wish to extend the detector to detect an ``apple'' class, and additionally detect the fine-grained type of animal, e.g.\ ``cat'', ``dog'', or ``elephant'' within the ``animal'' super class. Note that this setting is similar to the recently introduced incremental implicitly-refined classification~\cite{abdelsalam2021iirc} task.

We introduce a benchmark for the refined few-shot detection task based on the MS-COCO dataset.
We note that the 80 classes in MS-COCO are already divided into 11 super-categories~\cite{lin2014microsoft}.
Among these 11 super-categories, we choose the ``animal'' and ``food'' categories containing 10 child classes each as our super-base classes. Next, the 20 child classes belonging to the ``person'', ``accessory'', ``vehicle'', and ``furniture'' super-categories are selected as the novel classes.
The remaining 40 child classes constitute the standard base classes. Similar to the standard few-shot detection setting, the method is first provided with a large base dataset containing annotations for the 40 standard base classes, as well as the 2 super base classes, which can be used to learn the base model. Next, in the few-shot adaption stage, the method is given only $K$ training images containing annotations for each of the novel classes, as well as the child classes of the two base super-classes. Given this $K$-shot dataset, the method should learn to detect the novel classes as well as child classes of the base super-classes, while retaining its performance on the standard base classes. For each method, we report the combined AP over the 80 classes, as well as the performance over the 40 base classes (bAP), and 20 novel classes (nAP). We also report the average precision over child classes of the two super-classes, namely ``animal'' (aAP) and ``food'' (fAP). The results are aggregated over 10 random few-shot datasets. 

\begin{figure*}[ht]
    \centering%
    	\newcommand{\wid}{0.99\textwidth}%
    	\includegraphics*[trim = 0 200 0 0, width = \wid]{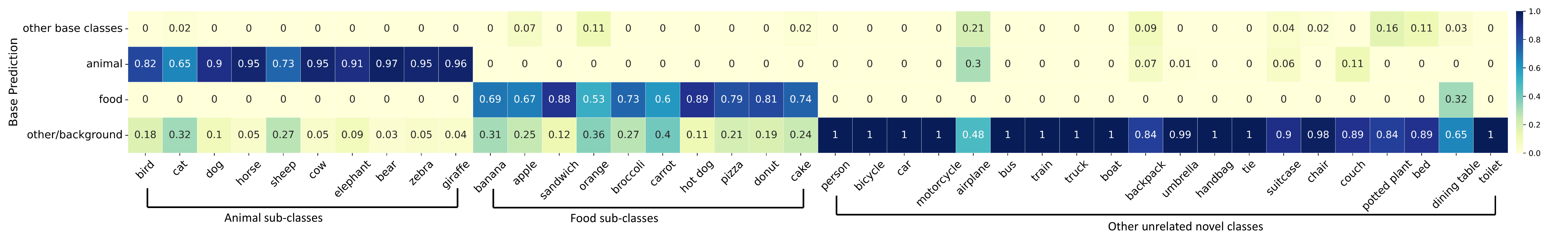}\vspace{-1mm}%
    	\captionof{figure}{Analysis of base model behaviour on novel classes. See Sec.\ \ref{exp:cls_refined} for details.
    	}%
    	\label{fig:base_model_analysis}%
    	\vspace{-5mm}%
\end{figure*}

\parsection{Results} We compare our approach HDA with the baseline TFA~\cite{wang2020frustratingly} with cosine classifier. For both methods, we first train a base model to detect the 42 classes in the base dataset, using the same hyper-parameters as in TFA's base training. In order to perform few-shot detection with TFA, we fine-tune a predictor $\mathcal{P}_n$ to make joint detection over all 80 classes. 
For our approach HDA, we use the class hierarchy provided in the COCO dataset to establish the relation between the novel classes and the existing base classes.
Table~\ref{tab:refined_fsdet} shows the results of this comparison over 1, 5, and 10 shot datasets. Our approach HDA outperforms the TFA baseline on the novel classes which are unrelated to the base classes (nAP). The fine-grained detection performance on the two super classes, namely ``animal'' (aAP) and ``food'' (fAP), is lower for HDA, compared to TFA. On the other hand, our approach obtains substantial improvement in the base class performance (bAP) over the baseline TFA. Consequently, in terms of the combined performance over all classes (AP), HDA outperforms TFA by over $3.7$ points for all shots. 

\parsection{Analysis of base detector} The performance of our approach relies on the ability of the base detector to correctly classify the novel classes into the correspondng super-class or background. We analyse the behaviour of our base detector in Fig.\ \ref{fig:base_model_analysis}. For each novel class (x-axis) in a 10-shot dataset for class-refined detection, we pass its ground truth proposals through the base model. We then plot the fraction of times the base model classifies a novel class object as background, animal or food super-class, or other base classes. 
We observe that in majority of the cases, the base model can correctly classify the different types of animals and food items as the animal and food super-class, respectively, while classifying the unrelated novel classes as background. However we note that the base model can struggle to classify certain novel classes (e.g.\ ``airplane'', ``orange'').

%% file: 05_conclusion.tex
\section{CONCLUSION}
We propose a hierarchical approach for few-shot detection which addresses the ``catastrophic forgetting'' on the base classes. The detection head for the novel classes is trained using a specialized optimization strategy, leading to significantly lower training times, compared to stochastic gradient descent. Our approach obtains competitive novel class performance on few-shot COCO benchmark, while retaining the performance of the base model on the base classes. We further demonstrate the application of our approach to a new class-refined few-shot detection task. 

\parsection{Limitation and future work} The performance of our approach relies on the availability of a strong base detector. Our approach will fail if the base detector cannot generate proposals for certain novel objects. Addressing this limitation by e.g.\ finetuning the base detector is an interesting future work. Extending our approach for incremental learning setting is another possible future work.


 

%% file: supp.tex
\setcounter{section}{0}
\renewcommand{\thesection}{\Roman{section}}

\begin{center}
	\textbf{\large Supplementary Material}
\end{center}

\input{06_supplementary}

%% file: 06_supplementary.tex
In this supplementary material, we provide additional details and analysis of our approach. First, we describe a variant of our optimization algorithm utilizing mini-batches in Section~\ref{sec:optim_mini_batch}. Additional implementation details about the training and inference of our approach is provided in Section~\ref{sec:impl_details}. Detailed results and runtime analysis is provided in Section~\ref{sec:detailed_results}

\section{Fast Optimization with mini-batches}
\label{sec:optim_mini_batch}
Our Newton's method based fast optimization strategy presented in Section \ref{sec: fast optimization for few-shot detection} in the main paper computes the update step \ref{eq:update_step} using all available data. While such a strategy is well suited to the few-shot detection scenario with a moderate number of classes, it suffers from GPU memory limitations if the number of classes is high, or the GPU memory available is low. Thus, here we present a variant of our optimization approach which uses mini-batches. Similar to stochastic gradient descent, we randomly sample a mini-batch with 16 images in each iteration and compute the weight updates using these images. 
Note that as the update step \ref{eq:update_step} is estimated using a subset of training samples, it can be noisy and hence lead to training instabilities. 
To avoid this, we add an additional $L_2$ regularization term to our quadratic approximation \ref{eq:l_quad_approx} to obtain,
\begin{equation}
\label{eq:l_quad_approx_reg}
    \Tilde{\mathcal{L}}_{w_i}(\Delta{w})\approx \mathcal{L}(w_i)+ \Delta w_i^T J_{w_i}+\frac{1}{2}\Delta w_i^T H_{w_i} \Delta w_i + \frac{\lambda}{2}\Delta w_i^T  \Delta w_i
\end{equation}
Here, $\lambda$ is the regularization factor. Consequently, the update step is obtained as,
\begin{equation}
\label{eq:update_step_reg}
    \Delta w_i = -(H_{w_i} + \lambda I)^{-1}\cdot J_{w_i}
\end{equation}
Here, $I$ is the identity matrix. The regularization term prevents instabilities in the inversion of the Hessian matrix $H_{w_i}$. Note that by using a high regularization factor, the update step \ref{eq:update_step_reg} will be approximately equal to the vanilla gradient descent step with a learning rate of $\frac{1}{\lambda}$. We set the regularization factor $\lambda = 0.5$ in all our experiments. 
For every batch, we run 1 Newton iteration with $N_{CG}=2$ CG iterations. 


\section{Additional Implementation Details}
\label{sec:impl_details}
In this section, we provide details about the weight initialization strategy used when training the novel predictor $\mathcal{P}_n$, as well as the proposal filtering strategy used in our approach HDA.

\subsection{Weight Initialization}
Here, we describe the strategy used to initialize the weights of the novel predictor $\mathcal{P}_n$ for our experiments in Section \ref{sec: experiments}.

\parsection{TFA$^*$} We follow the approach used by the original authors~\cite{wang2020frustratingly} when evaluating \texttt{TFA$^*$} on the MS-COCO based generalized few-shot detection benchmark in Section \ref{exp:comparion} and \ref{exp:ablation}. That is, we first pretrain the novel class weights on a few-shot dataset which only contains novel classes. The pretrained novel class weights are then concatenated with the weights of pretrained base predictor $\mathcal{P}_b$ to obtain the initial weights for $\mathcal{P}_n$. When evaluating on the class-refined few-shot detection benchmark in Section \ref{exp:cls_refined}, we initialize the classifier and bounding box regressor weights for the novel classes randomly, using a zero-mean Gaussian with standard deviation 0.01. The classifier weights for the child classes of existing base super-classes are also initialized randomly in an identical manner. For the bounding box regressor of the child classes, we assign the pretrained base predictor weights of corresponding base super-class as the initial weights. These novel and child class weights are then concatenated with the pretrained weights of the standard base classes to obtain the initial weights for the novel predictor $\mathcal{P}_n$



\parsection{TFA-Fast} For evaluation on the generalized few-shot detection benchmark in Section \ref{exp:comparion} and \ref{exp:ablation}, we initialize the weights for novel classes randomly using a zero-mean normal distribution with standard deviation 0.01. These weights are then concatenated with the weights of the pretrained base predictor $\mathcal{P}_b$ to obtain the initial weights for $\mathcal{P}_n$. 

\parsection{HDA} The base predictor of HDA have the weights same as the base model. For evaluation on the generalized few-shot detection benchmark in Section \ref{exp:comparion} and \ref{exp:ablation}, we initialize the weights for the novel predictor randomly using a zero-mean normal distribution with standard deviation 0.01.  For evaluation on the class-refined few-shot detection benchmark in Section \ref{exp:cls_refined}, we initialize the novel predictor weights, as well as the classifier weights for the child classes of existing base super-classes in an identical manner. For the bounding box regressor of the child classes, we assign the pre-trained base predictor weights of corresponding base super-class as the initial weights. 

\subsection{Proposal Filtering in HDA}
For both training and inference in HDA, we follow the same procedure of proposal filtering. The proposals and associated box features are first passed to the base predictor to do inference for base classes. The inference follows the standard process of Faster R-CNN and uses the same hyper-parameters as our TFA baseline, wherein the threshold for classification scores is 0.05, threshold for Non-Maximum Suppression is 0.5 and top 100 detections per image are output. Then from the remaining proposals that are not inferred as base detections, we select proposals and associated box features that have the maximum class scores of each super base class or ``other/background'' as the input for corresponding predictor head.

\section{Detailed Results}
\label{sec:detailed_results}

\subsection{Generalized few-shot detection}
In this section, we provide detailed results and runtime analysis on the generalized few-shot detection benchmark used in Section \ref{exp:comparion} and \ref{exp:ablation} in the main paper.

\parsection{Results} In Table \ref{tab:sota_std} of the main paper, we report the results of our approach on the MS-COCO generalized few-shot detection benchmark. Here, we provide further detailed results in Table~\ref{tab:appendix-genralized few-shot object detection}. We compare our approach HDA with the baselines TFA$^*$, TFA-Fast, and HDA-wo-Aug introduced in the main paper. Additionally, we also report results of a variant of our approach HDA-wo-Aug-mb employing the  optimization strategy with mini-batches introduced in Section~\ref{sec:optim_mini_batch}, without any data augmentation. For each method, we report the mean AP score, as well as the average precision scores at overlap thresholds 0.5 (AP50) and 0.75 (AP75) over all classes, base classes, and novel classes. For each metric, we report the average score as well as the 95$\%$ confidence interval computed over 10 random few-shot datasets. Compared to TFA$^*$, our approach obtains superior novel class performance (nAP), as well as overall AP, across all shots. Furthermore, note that our approach HDA-wo-Aug-mb employing mini-batches obtains results comparable to the standard HDA.

\parsection{Runtime comparison} We provide a detailed runtime comparison of our approach HDA with baseline TFA$^*$, as well as the variants TFA-Fast and HDA-wo-Aug introduced in the main paper in Section \ref{exp:ablation}. For each method, we report in Table~\ref{tab: time of t and n} the total time spent on feature extraction ($T_{feat}$), total time spent on optimization ($T_{optim}$), the time per optimization iteration ($t$) and the total number of iterations ($n$). For TFA$^*$, we report the time for both the pre-training of the novel class weights, as well as the joint finetuning of novel and base class weights. Our Newton's method based optimization strategy takes higher time per iteration compared to stochastic gradient descent (SGD) due to the computation of the Hessian matrix $H_{w_i}$ and the Conjugate Gradient iterations involved in the computation of the update step. 
However, only 30 iterations are sufficient to obtain good performance with our approach, compared to 80000 iterations needed by SGD for the 5-shot case. Consequently, our approach obtains over 200 times speedup in total optimization time, compared to SGD, on the 5 shot dataset. 

We also provide a runtime comparison between our approach HDA-wo-Aug-mb using mini-batches and TFA employing stochastic gradient descent. For both approaches, we use the same mini-batch size and extract the features in each iteration. The total runtime ($T$), the time per iteration ($t$), and the total number of iterations ($n$) are reported  in Table~\ref{tab: time of t and n of HDA+CG}. Our approach obtains over 150 times speedup compared to TFA in the 5-shot setting.

\begin{figure*}[h]
     \centering
         \includegraphics[width=\textwidth]{Figures/TFA+SGD_HDA+CG+aug_1-Shot_horizontal.png}
     \vspace{0pt}
         \includegraphics[width=\textwidth]{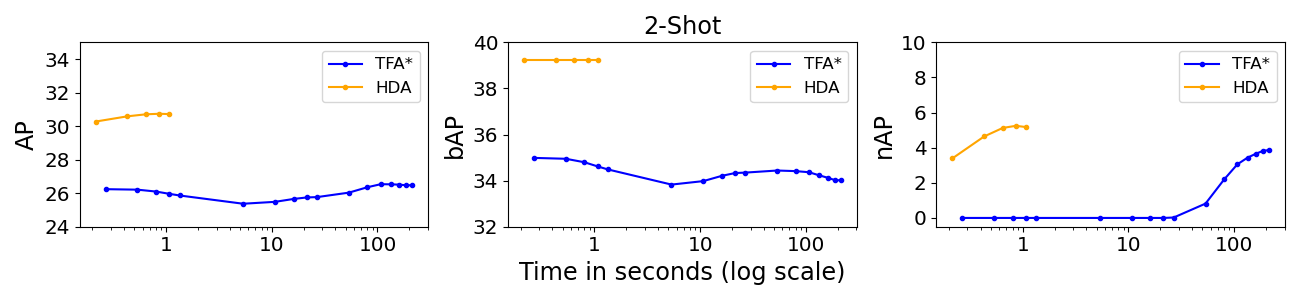}
     \vspace{0pt}
         \includegraphics[width=\textwidth]{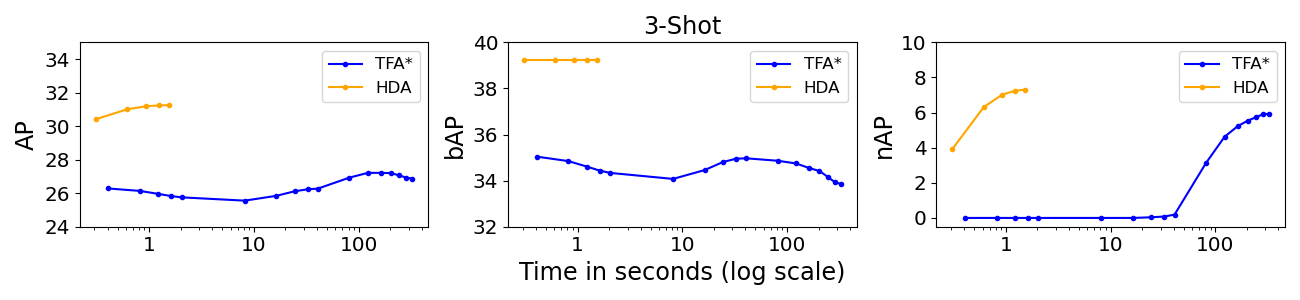}
     \vspace{0pt}
         \includegraphics[width=\textwidth]{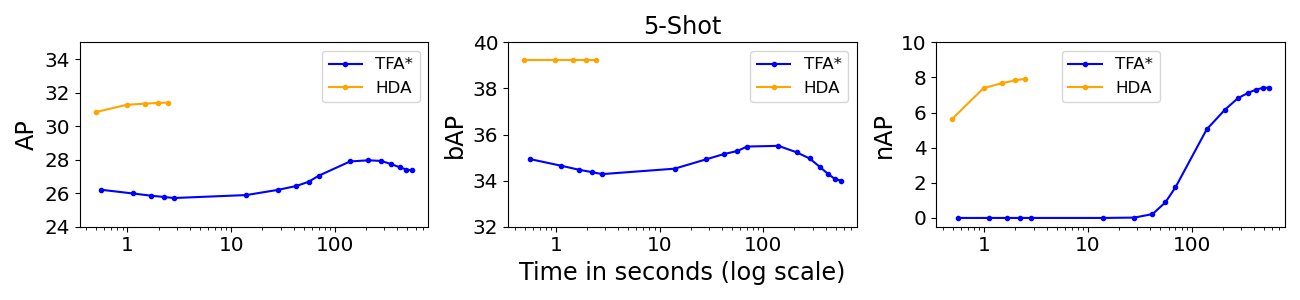}
     \vspace{0pt}
         \includegraphics[width=\textwidth]{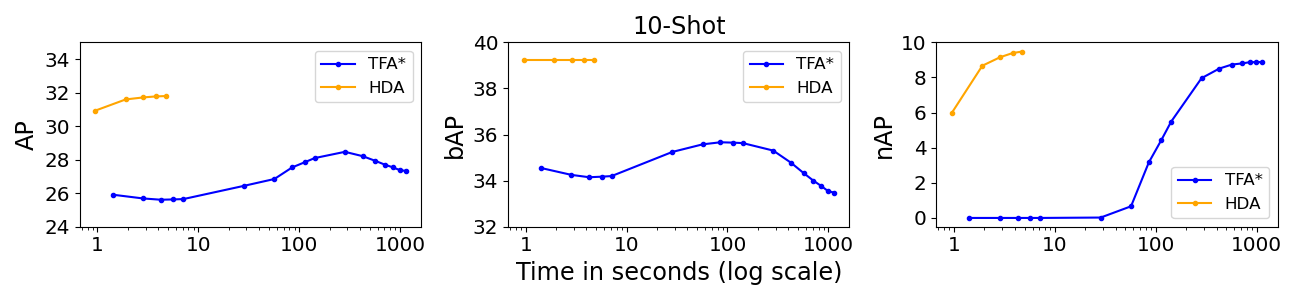}

     \vspace{0mm}
    \caption{Convergence plots showing the model performance, in terms of AP, bAP, and nAP, for different training times (in seconds), for 1, 2, 3, 5, and 10-shot datasets. We exclude the initial time for feature extraction. Note that the x-axis follows logarithmic scale.}\vspace{-3mm}
    \label{fig:convergence_plot}
\end{figure*}

\parsection{Convergence plots} In the main paper, we provide convergence plots comparing the performance of HDA and TFA$^*$ for different training time for the 1-shot dataset. Here, we additionally provide similar convergence plots for 2, 3, 5, and 10-shot datasets in Figure~\ref{fig:convergence_plot}. Our approach HDA obtains substantially convergence in terms of nAP score across all datasets.  

\subsection{Class-refined few-shot detection}
Here, we report detailed results on the class-refined few-shot detection benchmark introduced in Section \ref{exp:cls_refined} in the main paper. In Table~\ref{tab:appendx-class refined few-shot detection}, we report the performance in terms of AP, AP50, and AP75 over all classes, the novel classes, the standard base classes, as well as the child classes of the animal and food super-classes. For each metric, we report the average score as well as the 95$\%$ confidence interval computed over 10 random dataset splits. The results are reported for 1, 2, 3, 5, and 10-shot datasets. 

\begin{table*}[t] 
\centering
\resizebox{\textwidth}{!}{
    \begin{tabular}{l|l|ccc|ccc|ccc}
    \toprule
    \multirow{2}*{Shots} & \multirow{2}*{Method} & \multicolumn{3}{c|}{Overall} & \multicolumn{3}{c|}{Base class} & \multicolumn{3}{c}{Novel class} \\ 
    \cmidrule(lr){3-5} \cmidrule(lr){6-8} \cmidrule(lr){9-11}
    ~&~& AP & AP50 & AP75 & bAP & bAP50 & bAP75 & nAP & nAP50 & nAP75  \\ 
    \midrule
    \multirow{5}*{1}& HDA &30.2$\pm$0.1 &46.2$\pm$0.2 &32.7$\pm$0.1 &39.2$\pm$0.0 &59.3$\pm$0.0 &42.8$\pm$0.0 &3.0$\pm$0.4 &6.8$\pm$0.7 &2.3$\pm$0.5  \\
    ~& HDA-wo-Aug &30.0$\pm$0.1 &45.8$\pm$0.2 &32.6$\pm$0.1 &39.2$\pm$0.0 &59.3$\pm$0.0 &42.8$\pm$0.0 &2.4$\pm$0.3 &5.2$\pm$0.6 &1.9$\pm$0.4  \\
    ~& HDA-wo-Aug-mb &30.1$\pm$0.1 &45.9$\pm$0.2 &32.7$\pm$0.1 &39.2$\pm$0.0 &59.3$\pm$0.0 &42.8$\pm$0.0 &2.9$\pm$0.4 &5.8$\pm$0.7 &2.5$\pm$0.5  \\
    ~& TFA-Fast  &24.2$\pm$0.6 &40.5$\pm$0.6 &25.3$\pm$1.0 &31.4$\pm$0.7 &52.2$\pm$0.6 &33.0$\pm$1.2 &2.6$\pm$0.4 &5.4$\pm$0.5 &2.3$\pm$0.5 \\
    ~& TFA$^*$ &24.1$\pm$0.5 &40.3$\pm$0.5 &25.4$\pm$0.9 &31.4$\pm$0.6 &52.2$\pm$0.6 &33.2$\pm$1.0 &2.3$\pm$0.3 &4.4$\pm$0.4 &2.1$\pm$0.4 \\
    \midrule
    \multirow{4}*{2}& HDA &30.6$\pm$0.1 &47.1$\pm$0.2 &33.0$\pm$0.1 &39.2$\pm$0.0 &59.3$\pm$0.0 &42.8$\pm$0.0 &4.6$\pm$0.4 &10.4$\pm$0.8 &3.6$\pm$0.5  \\
    ~& HDA-wo-Aug &30.4$\pm$0.1 &46.6$\pm$0.2 &32.9$\pm$0.1 &39.2$\pm$0.0 &59.3$\pm$0.0 &42.8$\pm$0.0 &4.1$\pm$0.4 &8.7$\pm$0.7 &3.4$\pm$0.3  \\
    ~& HDA-wo-Aug-mb &30.6$\pm$0.1 &46.8$\pm$0.2 &33.1$\pm$0.1 &39.2$\pm$0.0 &59.3$\pm$0.0 &42.8$\pm$0.0 &4.5$\pm$0.4 &9.2$\pm$0.7 &4.0$\pm$0.4 \\
    ~& TFA-Fast &25.3$\pm$0.5 &41.8$\pm$0.6 &27.0$\pm$0.6 &32.4$\pm$0.6 &53.0$\pm$0.7 &34.9$\pm$0.8 &3.9$\pm$0.4 &8.1$\pm$0.9 &3.5$\pm$0.4   \\
    ~& TFA$^*$ &24.7$\pm$0.6 &40.5$\pm$0.7 &26.4$\pm$0.8 &31.6$\pm$0.7 &51.5$\pm$0.8 &34.1$\pm$0.9 &3.8$\pm$0.4 &7.5$\pm$0.7 &3.4$\pm$0.4 \\
    \midrule
    \multirow{5}*{3}& HDA &30.8$\pm$0.1 &47.5$\pm$0.2 &33.2$\pm$0.1 &39.2$\pm$0.0 &59.3$\pm$0.0 &42.8$\pm$0.0 &5.6$\pm$0.4 &12.0$\pm$0.8 &4.6$\pm$0.4  \\
    ~& HDA-wo-Aug &30.7$\pm$0.1 &47.1$\pm$0.3 &33.2$\pm$0.1 &39.2$\pm$0.0 &59.3$\pm$0.0 &42.8$\pm$0.0 &5.1$\pm$0.5 &10.3$\pm$1.0 &4.5$\pm$0.5  \\
    ~& HDA-wo-Aug-mb &30.8$\pm$0.1 &47.1$\pm$0.3 &33.3$\pm$0.1 &39.2$\pm$0.0 &59.3$\pm$0.0 &42.8$\pm$0.0 &5.4$\pm$0.6 &10.6$\pm$1.1 &4.9$\pm$0.6 \\
    ~& TFA-Fast &26.2$\pm$0.4 &42.7$\pm$0.6 &28.3$\pm$0.5 &33.3$\pm$0.4 &53.7$\pm$0.5 &36.2$\pm$0.6 &4.9$\pm$0.5 &9.8$\pm$1.1 &4.5$\pm$0.5   \\
    ~& TFA$^*$ &25.0$\pm$0.6 &40.6$\pm$0.8 &27.0$\pm$0.6 &31.6$\pm$0.6 &51.0$\pm$0.9 &34.5$\pm$0.7 &4.9$\pm$0.6 &9.4$\pm$1.1 &4.6$\pm$0.6 \\
    \midrule
    \multirow{5}*{5}& HDA &31.2$\pm$0.1 &48.2$\pm$0.2 &33.6$\pm$0.1 &39.2$\pm$0.0 &59.3$\pm$0.0 &42.8$\pm$0.0 &7.1$\pm$0.5 &14.9$\pm$0.9 &6.0$\pm$0.6 \\
    ~& HDA-wo-Aug &31.0$\pm$0.2 &47.7$\pm$0.3 &33.5$\pm$0.2 &39.2$\pm$0.0 &59.3$\pm$0.0 &42.8$\pm$0.0 &6.4$\pm$0.6 &12.8$\pm$1.1 &5.8$\pm$0.7  \\
    & HDA-wo-Aug-mb &31.1$\pm$0.2 &47.8$\pm$0.3 &33.6$\pm$0.2 &39.2$\pm$0.0 &59.3$\pm$0.0 &42.8$\pm$0.0 &6.8$\pm$0.6 &13.3$\pm$1.0 &6.2$\pm$0.7  \\
    ~& TFA-Fast  &27.3$\pm$0.3 &44.1$\pm$0.5 &29.7$\pm$0.4 &34.3$\pm$0.3 &54.6$\pm$0.4 &37.6$\pm$0.3 &6.4$\pm$0.5 &12.5$\pm$1.0 &6.0$\pm$0.6  \\
    ~& TFA$^*$ &25.4$\pm$0.6 &41.3$\pm$0.8 &27.6$\pm$0.6 &31.8$\pm$0.6 &50.9$\pm$0.7 &34.8$\pm$0.6 &6.4$\pm$0.6 &12.6$\pm$1.1 &6.0$\pm$0.7  \\
    \midrule
    \multirow{5}*{10}& HDA &31.7$\pm$0.1 &49.1$\pm$0.2 &34.1$\pm$0.1 &39.2$\pm$0.0 &59.3$\pm$0.0 &42.8$\pm$0.0 &9.0$\pm$0.4 &18.3$\pm$0.9 &7.9$\pm$0.4  \\
    ~& HDA-wo-Aug &31.4$\pm$0.2 &48.3$\pm$0.3 &34.0$\pm$0.2 &39.2$\pm$0.0 &59.3$\pm$0.0 &42.8$\pm$0.0 &7.9$\pm$0.6 &15.2$\pm$1.2 &7.5$\pm$0.6  \\
    ~& HDA-wo-Aug-mb &31.6$\pm$0.2 &48.6$\pm$0.3 &34.1$\pm$0.2 &39.2$\pm$0.0 &59.3$\pm$0.0 &42.8$\pm$0.0 &8.6$\pm$0.6 &16.5$\pm$1.2 &8.2$\pm$0.6  \\
    ~& TFA-Fast &28.5$\pm$0.3 &45.6$\pm$0.4 &31.0$\pm$0.2 &35.2$\pm$0.2 &55.4$\pm$0.2 &38.6$\pm$0.2 &8.4$\pm$0.6 &16.1$\pm$1.2 &8.1$\pm$0.5   \\
    ~& TFA$^*$ &25.8$\pm$0.4 &42.3$\pm$0.7 &27.8$\pm$0.4 &31.6$\pm$0.4 &50.9$\pm$0.6 &34.5$\pm$0.5 &8.3$\pm$0.5 &16.4$\pm$1.0 &7.7$\pm$0.5 \\
    \bottomrule
    \end{tabular}
}
\caption{Generalized few-shot detection performance on MS-COCO~\cite{lin2014microsoft} dataset. We report the mean AP score, as well as the average precision scores at overlap thresholds 0.5 (AP50) and 0.75 (AP75) over all classes (overall), base classes, and novel classes. For each metric, we report the average score as well as the 95$\%$ confidence interval computed over 10 random few-shot datasets.}
\label{tab:appendix-genralized few-shot object detection}
\end{table*}

\begin{table*}[t] 
\centering
\resizebox{\textwidth}{!}{
\begin{tabular}{l|ccc|ccc|ccc|ccc}
\toprule
\multirow{2}*{Model} & \multicolumn{3}{c|}{$T_{feat}$ ($s$)} & \multicolumn{3}{c|}{$T_{optim}$ ($s$)} & \multicolumn{3}{c|}{$t$ ($s$)} & \multicolumn{3}{c}{$n$} \\ 
\cmidrule(lr){2-4} \cmidrule(lr){5-7} \cmidrule(lr){8-10} \cmidrule{11-13}
~& 1 & 5 & 10 & 1 & 5 & 10 & 1 & 5 & 10 & 1 & 5 & 10 \\ 
\midrule
\rowcolor{Gray}
HDA   & 9.56 & 43.27 & 89.41 & 0.63 & 2.47 & 4.77 & 0.0211 & 0.0822 & 0.1589 & 30 & 30 & 30 \\ 
\rowcolor{Gray}
HDA-wo-Aug        & 9.75 & 44.53 & 89.97 & 0.57 & 2.21 & 4.22 & 0.0191 & 0.0737 & 0.1407 & 30 & 30 & 30 \\
TFA-Fast        & 9.36 & 37.53 & 74.42 & 3.78 & 16.83 & 33.39 & 0.0378 & 0.1683 & 0.3339 & 100 & 100 & 100  \\
TFA$^*$ (pretraining)      & 3.03 & 10.38 & 19.58 & 2.75 & 8.25 & 11.60 & 0.0055 & 0.0055 & 0.0058 & 500 & 1500 & 2000 \\
TFA$^*$ (finetuning)      & 8.55 & 39.20 & 78.76 & 107.20 & 560.00 & 1136.00 & 0.0067 &0.0070 & 0.0071 & 16000 & 80000 & 160000 \\
\bottomrule
\end{tabular}
}
\caption{Runtime analysis on the generalized few-shot detection benchmark. For each method, we report the total time for feature extraction ($T_{feat}$), total time for optimization ($T_{optim}$), time per optimization iteration ($t$) and the total number of optimization iterations ($n$). All times are reported in seconds.}
\label{tab: time of t and n}
\end{table*}

\begin{table*}[t] 
\centering
\resizebox{\textwidth}{!}{
\begin{tabular}{l|ccc|ccc|ccc}
\toprule
\multirow{2}*{Model} & \multicolumn{3}{c|}{$T$ ($s$)} & \multicolumn{3}{c|}{$t$ ($s$)} & \multicolumn{3}{c}{$n$} \\ 
\cmidrule(lr){2-4} \cmidrule(lr){5-7} \cmidrule(lr){8-10} 
~& 1 & 5 & 10 & 1 & 5 & 10 & 1 & 5 & 10 \\ 
\midrule
\rowcolor{Gray}
HDA-wo-Aug-mb  & 413.56 & 672.51 & 882.00 & 2.0678 &2.2417 & 2.2050 & 200 & 300 & 400 \\ 
TFA & 21041.60 & 105832.00 & 211440.00 & 1.3151 & 1.3229 & 1.3215 & 16000 & 80000 & 160000 \\
\bottomrule
\end{tabular}
}
\caption{Runtime comparison between HDA-wo-Aug-mb employing our fast optimization strategy with mini-batches, and original TFA employing stochastic gradient descent. We report the total runtime ($T$), and the time per optimization iteration ($t$) including the time take to extract the mini-batch. We also report the total number of iterations ($n$) employed for each method. All times are in seconds.} 
\label{tab: time of t and n of HDA+CG}
\end{table*}

\begin{table*}[t] 
\centering
\resizebox{\textwidth}{!}{
    \begin{tabular}{l|l|ccc|ccc|ccc|ccc|ccc}
    \toprule
    \multirow{2}*{Shots} & \multirow{2}*{Method} & \multicolumn{3}{c|}{Overall} & \multicolumn{3}{c|}{Base class} & \multicolumn{3}{c|}{Animal super class} & \multicolumn{3}{c|}{Food super class} & \multicolumn{3}{c}{Novel class} \\ 
    \cmidrule(lr){3-5} \cmidrule(lr){6-8} \cmidrule(lr){9-11} \cmidrule(lr){12-14} \cmidrule(lr){15-17}
    ~&~& AP & AP50 & AP75 & bAP & bAP50 & bAP75 & aAP & aAP50 & aAP75 & fAP & fAP50 & fAP75 & nAP & nAP50 & nAP75  \\ 
    \midrule
    \multirow{3}*{1}& HDA &21.2$\pm$0.1 &32.9$\pm$0.1 &23.0$\pm$0.2 &39.2$\pm$0.0 &59.8$\pm$0.0 &43.0$\pm$0.0 &6.9$\pm$0.5 &12.3$\pm$0.8 &7.1$\pm$0.7 &4.0$\pm$0.7 &6.7$\pm$1.0 &4.2$\pm$0.8 &0.9$\pm$0.2 &2.6$\pm$0.4 &0.4$\pm$0.1  \\
    ~& HDA-wo-Aug &21.2$\pm$0.1 &32.9$\pm$0.2 &23.1$\pm$0.1 &39.2$\pm$0.0 &59.8$\pm$0.0 &43.0$\pm$0.0 &7.3$\pm$0.5 &12.3$\pm$0.8 &7.6$\pm$0.6 &4.1$\pm$0.6 &6.8$\pm$1.0 &4.4$\pm$0.7 &0.8$\pm$0.2 &2.2$\pm$0.4 &0.5$\pm$0.1  \\
    ~& TFA &16.9$\pm$0.3 &28.5$\pm$0.4 &17.9$\pm$0.5 &30.6$\pm$0.6 &52.0$\pm$0.7 &32.2$\pm$1.1 &7.7$\pm$0.6 &11.5$\pm$0.8 &8.8$\pm$0.7 &4.7$\pm$0.8 &6.6$\pm$1.2 &5.4$\pm$1.0 &0.4$\pm$0.2 &0.9$\pm$0.4 &0.3$\pm$0.2  \\
    \midrule
    \multirow{3}*{2}& HDA &21.7$\pm$0.1 &33.9$\pm$0.1 &23.5$\pm$0.1 &39.2$\pm$0.0 &59.8$\pm$0.0 &43.0$\pm$0.0 &8.3$\pm$0.7 &14.1$\pm$1.3 &8.6$\pm$0.7 &5.2$\pm$0.5 &8.7$\pm$0.8 &5.5$\pm$0.6 &1.7$\pm$0.2 &4.5$\pm$0.5 &1.0$\pm$0.2  \\
    ~& HDA-wo-Aug &21.7$\pm$0.1 &33.8$\pm$0.2 &23.6$\pm$0.1 &39.2$\pm$0.0 &59.8$\pm$0.0 &43.0$\pm$0.0 &8.6$\pm$0.7 &14.2$\pm$1.2 &9.1$\pm$0.8 &5.2$\pm$0.4 &8.5$\pm$0.7 &5.5$\pm$0.5 &1.6$\pm$0.2 &4.0$\pm$0.5 &1.0$\pm$0.2  \\
    ~& TFA &17.6$\pm$0.3 &28.8$\pm$0.5 &19.1$\pm$0.3 &30.8$\pm$0.6 &50.9$\pm$1.0 &33.4$\pm$0.7 &9.5$\pm$0.6 &13.9$\pm$1.0 &10.8$\pm$0.6 &5.9$\pm$0.4 &8.4$\pm$0.5 &6.6$\pm$0.4 &0.9$\pm$0.1 &2.0$\pm$0.2 &0.7$\pm$0.2  \\
    \midrule
    \multirow{3}*{3}& HDA &22.0$\pm$0.1 &34.4$\pm$0.1 &23.7$\pm$0.1 &39.2$\pm$0.0 &59.8$\pm$0.0 &43.0$\pm$0.0 &8.9$\pm$0.6 &15.1$\pm$1.1 &9.5$\pm$0.7 &5.6$\pm$0.7 &9.3$\pm$1.0 &6.0$\pm$0.7 &2.1$\pm$0.2 &5.5$\pm$0.3 &1.3$\pm$0.2  \\
    ~& HDA-wo-Aug &21.9$\pm$0.1 &34.2$\pm$0.2 &23.8$\pm$0.1 &39.2$\pm$0.0 &59.8$\pm$0.0 &43.0$\pm$0.0 &9.1$\pm$0.6 &14.9$\pm$1.1 &9.8$\pm$0.7 &5.6$\pm$0.6 &9.3$\pm$0.9 &5.9$\pm$0.6 &2.0$\pm$0.2 &5.0$\pm$0.3 &1.3$\pm$0.2  \\
    ~& TFA &18.0$\pm$0.4 &29.1$\pm$0.6 &19.7$\pm$0.4 &30.9$\pm$0.7 &50.4$\pm$1.0 &33.9$\pm$0.8 &10.7$\pm$0.7 &15.5$\pm$1.0 &12.2$\pm$0.9 &6.5$\pm$0.4 &9.4$\pm$0.6 &7.3$\pm$0.4 &1.5$\pm$0.2 &3.2$\pm$0.4 &1.3$\pm$0.2  \\
    \midrule
    \multirow{3}*{5}& HDA &22.4$\pm$0.1 &35.2$\pm$0.1 &24.1$\pm$0.1 &39.2$\pm$0.0 &59.8$\pm$0.0 &43.0$\pm$0.0 &10.5$\pm$0.6 &17.9$\pm$1.0 &10.9$\pm$0.7 &6.4$\pm$0.5 &10.7$\pm$0.8 &6.7$\pm$0.6 &2.7$\pm$0.2 &7.0$\pm$0.5 &1.7$\pm$0.1  \\
    ~& HDA-wo-Aug &22.4$\pm$0.1 &35.2$\pm$0.2 &24.2$\pm$0.1 &39.2$\pm$0.0 &59.8$\pm$0.0 &43.0$\pm$0.0 &10.7$\pm$0.6 &17.8$\pm$1.0 &11.3$\pm$0.7 &6.6$\pm$0.6 &10.9$\pm$0.8 &7.0$\pm$0.7 &2.6$\pm$0.2 &6.6$\pm$0.5 &1.7$\pm$0.1  \\
    ~& TFA &18.4$\pm$0.3 &29.9$\pm$0.5 &20.2$\pm$0.3 &30.9$\pm$0.6 &50.2$\pm$0.9 &33.8$\pm$0.6 &12.3$\pm$0.7 &18.0$\pm$1.2 &14.0$\pm$0.8 &7.5$\pm$0.5 &10.9$\pm$0.7 &8.4$\pm$0.5 &2.2$\pm$0.2 &4.6$\pm$0.4 &1.8$\pm$0.2  \\
    \midrule
    \multirow{3}*{10}& HDA &22.9$\pm$0.1 &36.2$\pm$0.1 &24.6$\pm$0.1 &39.2$\pm$0.0 &59.8$\pm$0.0 &43.0$\pm$0.0 &12.0$\pm$0.5 &20.3$\pm$0.9 &12.6$\pm$0.5 &7.6$\pm$0.5 &12.7$\pm$0.8 &8.0$\pm$0.6 &3.4$\pm$0.2 &8.4$\pm$0.4 &2.2$\pm$0.2  \\
    ~& HDA-wo-Aug &22.9$\pm$0.1 &36.1$\pm$0.1 &24.7$\pm$0.1 &39.2$\pm$0.0 &59.8$\pm$0.0 &43.0$\pm$0.0 &12.2$\pm$0.6 &20.5$\pm$1.0 &13.0$\pm$0.6 &7.5$\pm$0.5 &12.8$\pm$0.6 &7.8$\pm$0.7 &3.4$\pm$0.1 &8.1$\pm$0.3 &2.4$\pm$0.2  \\
    ~& TFA &19.2$\pm$0.2 &31.2$\pm$0.4 &20.9$\pm$0.2 &31.2$\pm$0.5 &50.6$\pm$0.8 &34.2$\pm$0.6 &14.2$\pm$0.8 &21.0$\pm$1.2 &16.2$\pm$1.0 &8.4$\pm$0.4 &12.7$\pm$0.5 &9.3$\pm$0.4 &3.1$\pm$0.1 &6.5$\pm$0.4 &2.5$\pm$0.2  \\
    \bottomrule
    \end{tabular}
}\vspace{-1mm}
\caption{Class-refined few-shot detection performance on MS-COCO~\cite{lin2014microsoft} dataset.}\vspace{-4mm}
\label{tab:appendx-class refined few-shot detection}
\end{table*}